\documentclass[journal,twoside]{IEEEtran}
\usepackage{cite}
\usepackage{amsmath,amssymb,amsfonts}
\usepackage{algorithmic}
\usepackage{graphicx}
\usepackage{textcomp}
\usepackage{stfloats}    
\usepackage{algorithm}      
\begin{document}
%
\title{Synchronous Unsupervised STDP Learning with Stochastic STT-MRAM Switching}
\author{
\IEEEauthorblockN{
Peng Zhou\IEEEauthorrefmark{1},
Julie A. Smith\IEEEauthorrefmark{1},
Laura Deremo\IEEEauthorrefmark{2},
Stephen K. Heinrich-Barna\IEEEauthorrefmark{2},
Joseph S. Friedman\IEEEauthorrefmark{1}
}\\
\IEEEauthorblockA{\IEEEauthorrefmark{1}The University of Texas at Dallas, email: joseph.friedman@utdallas.edu}\\
\IEEEauthorblockA{\IEEEauthorrefmark{2}Texas Instruments Inc.}  
}
\maketitle

\begin{abstract}
The use of analog resistance states for storing weights in neuromorphic systems is impeded by fabrication imprecision and device stochasticity that limit the precision of synapse weights. This challenge can be resolved by emulating analog behavior with the stochastic switching of the binary states of spin-transfer torque magnetoresistive random-access memory (STT-MRAM). However, previous approaches based on STT-MRAM operate in an asynchronous manner that is difficult to implement experimentally. This paper proposes a synchronous spiking neural network system with clocked circuits that perform unsupervised learning leveraging the stochastic switching of STT-MRAM. The proposed system enables a single-layer network to achieve 90\% inference accuracy on the MNIST dataset.
\end{abstract}

\begin{IEEEkeywords}
Spiking neural network; STT-MRAM; unsupervised learning; stochastic switching; spike-timing-dependent plasticity
\end{IEEEkeywords}

%
\IEEEpeerreviewmaketitle

\section{Introduction}
\begin{figure}[!t]
	\centering
	\includegraphics[width=1\columnwidth]{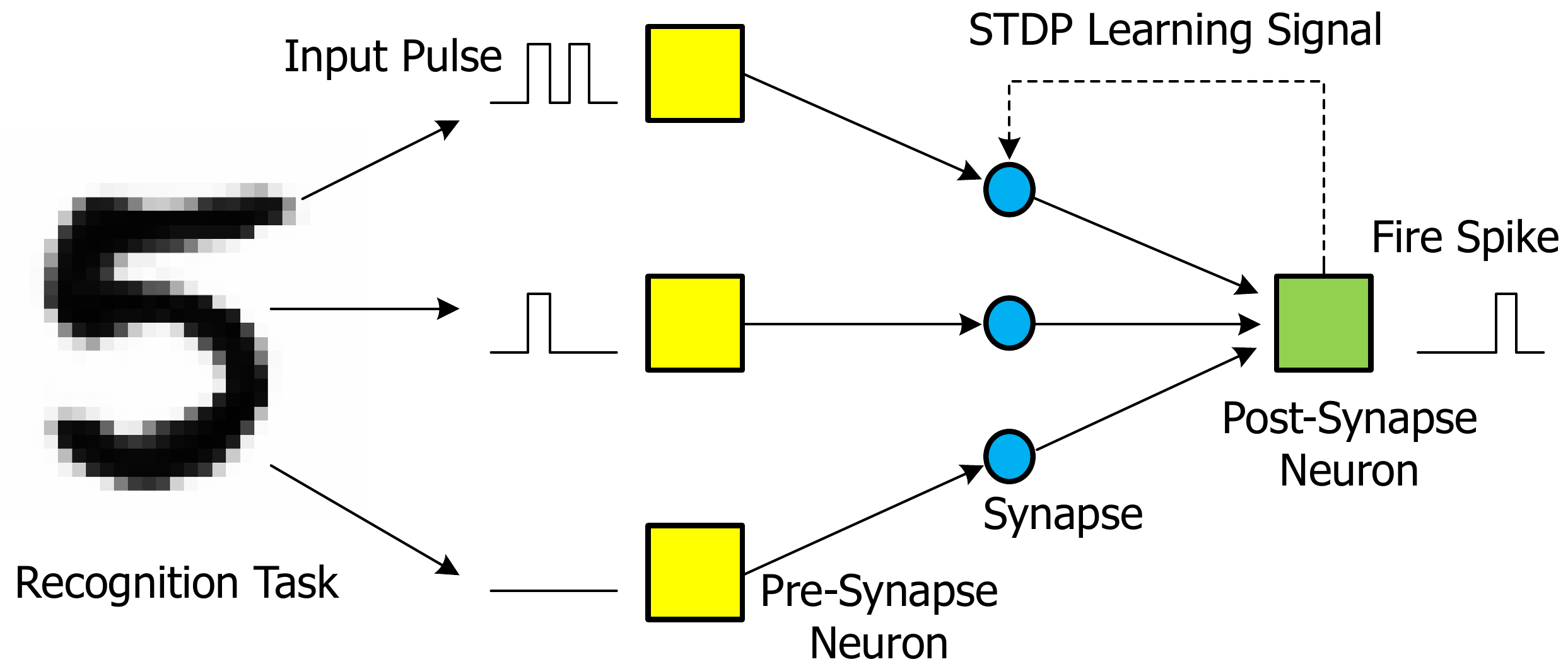}
	\caption{The SNN performs feed-forward recognition and feedback learning with an STDP learning rule.}
	\label{fig:fig1}
\end{figure}

\IEEEPARstart{S}{piking} neural networks (SNNs) can be used in energy-efficient neuromorphic computing systems that process information from spikes or pulses emitted by artificial neurons, as shown in Fig. \ref{fig:fig1}\cite{merolla2014million}. The spiking signals flow through artificial synapses, in which the synapse weights can be efficiently stored via the resistances of non-volatile memory devices. These neuromorphic systems readily perform the critical neural network function of vector-matrix multiplication by applying an input vector of voltages to a non-volatile synapse crossbar array. These non-volatile devices conventionally store weights in an analog manner, and the inference accuracy of such neuromorphic systems is determined by the degree to which the analog resistance state can be precisely written and stored.

However, it is extremely difficult to precisely write and store analog resistance states in non-volatile memory devices, thereby threatening the primary advantages arising from their use in neuromorphic SNNs. In particular, while memristors and phase change memory (PCM) are widely used due to their non-volatility and analog resistances \cite{cai2019fully, sebastian2018tutorial}, modifications to the stored resistance state requires the intrinsically destructive processes of ion migration or modification of the crystalline structure\cite{chen2013endurance, kim2019phase}. These processes are thermally-dependent and therefore intrinsically stochastic, thereby preventing the precise writing of analog resistance states even with careful control of the write pulse voltage and duration\cite{ambrogio2014statistical}. Furthermore, the resistance of memristor or PCM synapses can drift over time, thereby modifying the weights stored in the array \cite{yu2011investigating}. In concert with the imprecision and variation inevitable when fabricating large memory arrays, these issues significantly degrade the neural network recognition accuracy and are inherent to non-volatile memory devices with analog resistances.

These challenges can be overcome by exploiting the analog stochastic switching and stable non-volatile binary storage of spin-transfer torque magnetoresistive random-access memory (STT-MRAM). The switching between the two STT-MRAM magnetic states is intrinsically stochastic, with a switching probability dependent on the write pulse voltage and duration time; while this is not ideal for conventional memory applications, it provides opportunities for neuromorphic computing \cite{devolder2008single}. Vincent \textit{et al.} suggested that the stochastic switching of STT-MRAM can emulate analog synapse behavior in a neuromorphic system, and proposed a rule for unsupervised spike-timing-dependent plasticity (STDP) learning\cite{vincent2015spin}. However, their hardware learning rule and time-domain implementation cannot be directly applied to practical circuits, as it does not fully consider the electrical behavior.

In this work, we therefore propose and design a complete neuromorphic system with an electrically-realistic hardware learning circuit that stochastically switches the STT-MRAM synapses. We demonstrate that this synchronous system performs STDP learning using clocked neuronal activity, and that it enables efficient on-chip unsupervised online learning and recognition.

\section{Background}

STT-MRAM provides non-volatile binary states with high endurance, while its stochastic switching provides the analog behavior necessary for neuromorphic computing. STT-MRAM therefore provides an opportunity for efficient and reliable neuromorphic computing systems that resolve the challenges faced by non-volatile devices with analog resistance states.

\subsection{Magnetoresistive Random-Access Memory (MRAM)} 

\begin{figure}[!t]
	\centering
	\includegraphics[width=1\columnwidth]{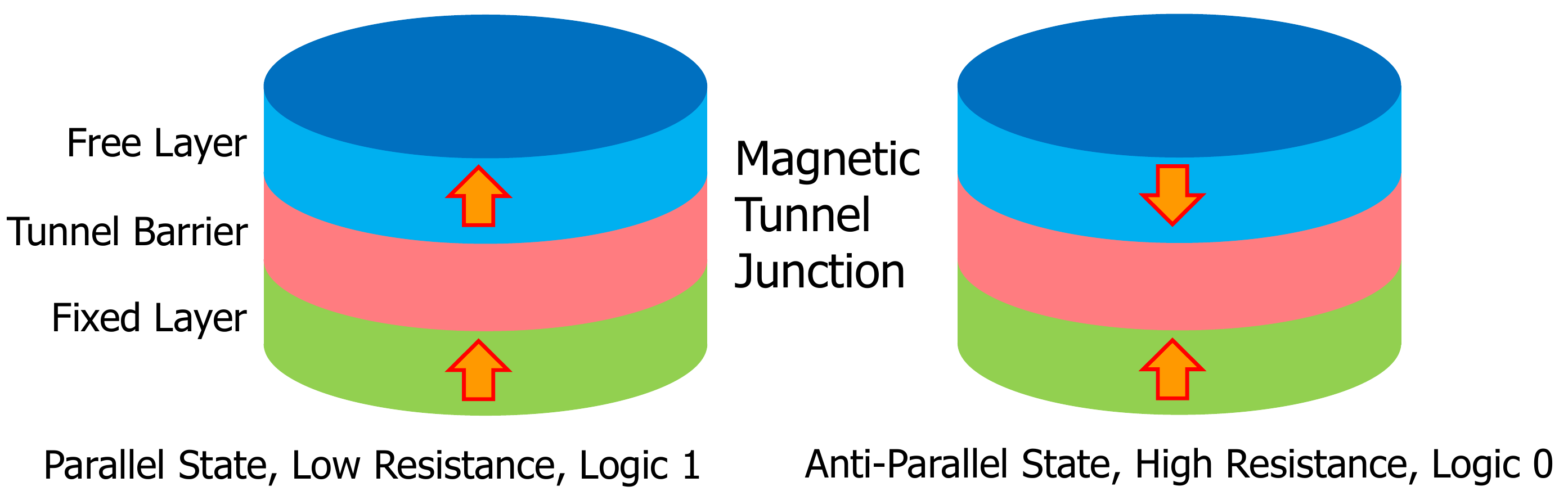}
	\caption{Structure of a magnetic tunnel junction.}
	\label{fig:fig2}
\end{figure}

The core component of an MRAM device is a magnetic tunnel junction (MTJ) which, as shown in Fig. \ref{fig:fig2}, consists of three layers: a fixed layer, a tunnel barrier, and a free layer. The fixed layer maintains a particular magnetic state, while the free layer switches between two magnetic states to produce binary resistance values. When the fixed and free layers have magnetizations in the same direction, the MTJ is in the parallel state and has low resistance; when the two magnetizations are in opposite directions, the MTJ is in the anti-parallel state and has a high resistance. These binary resistance states can represent synapse weights within a neuromorphic computing system.

\begin{figure}[!t]
	\centering
	\includegraphics[width=0.9\columnwidth]{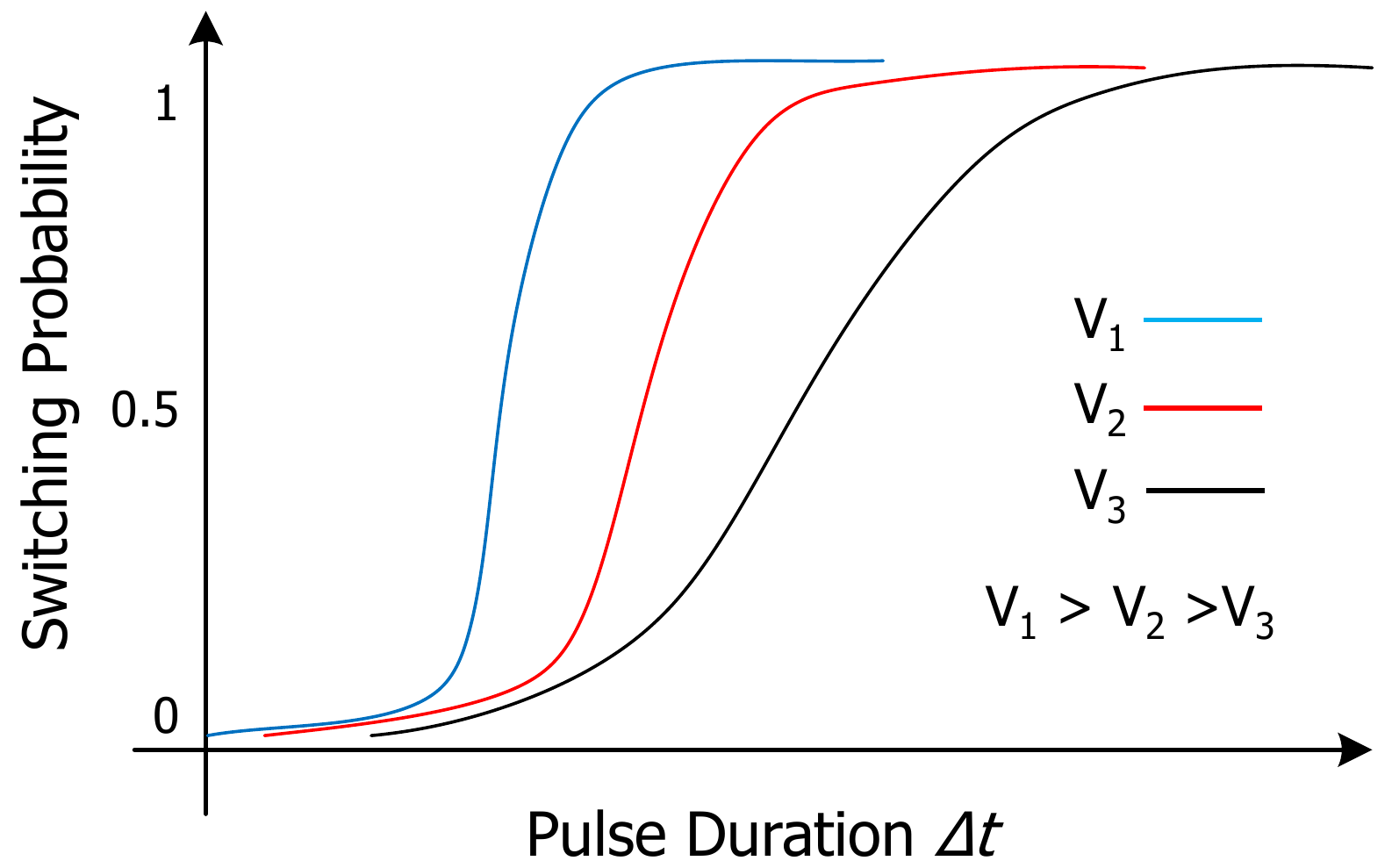}
	\caption{Stochastic switching of a magnetic tunnel junction.}
	\label{fig:fig3}
\end{figure}

STT is a stochastic process that switches the MTJ state within an STT-MRAM device. As shown in Fig. \ref{fig:fig3}, the switching probability increases with increasing pulse voltage and duration. This is a challenge for memory, as there is a relatively large minimum pulse voltage and duration to provide a sufficiently-high probability of MTJ switching. As described in section II-B, this stochastic STT-MRAM switching can be used to emulate analog synaptic behavior with binary MTJ resistance values.

The binary states of MRAM devices are highly stable and easy to differentiate, unlike the analog resistance states of memristors and PCM. While memristors, PCM, and MRAM all have switching mechanisms characterized by stochasticity, the binary nature of MRAM facilitates the writing into one of the two states; in contrast, memristors and PCM have numerous nearby states amongst which it is difficult to write precisely and to differentiate. This binary nature also provides MRAM with greater stability, as a high magnetic anisotropy prevents unwanted switching, whereas thermally-driven changes to memristor and PCM states can cause drift in their resistance. Furthermore, the ion motion and changes to crystalline structure underlying memristor and PCM switching, respectively, lead to significantly less endurance than the MRAM spin flipping.

\begin{figure}[!t]
	\centering
	\includegraphics[width=0.9\columnwidth]{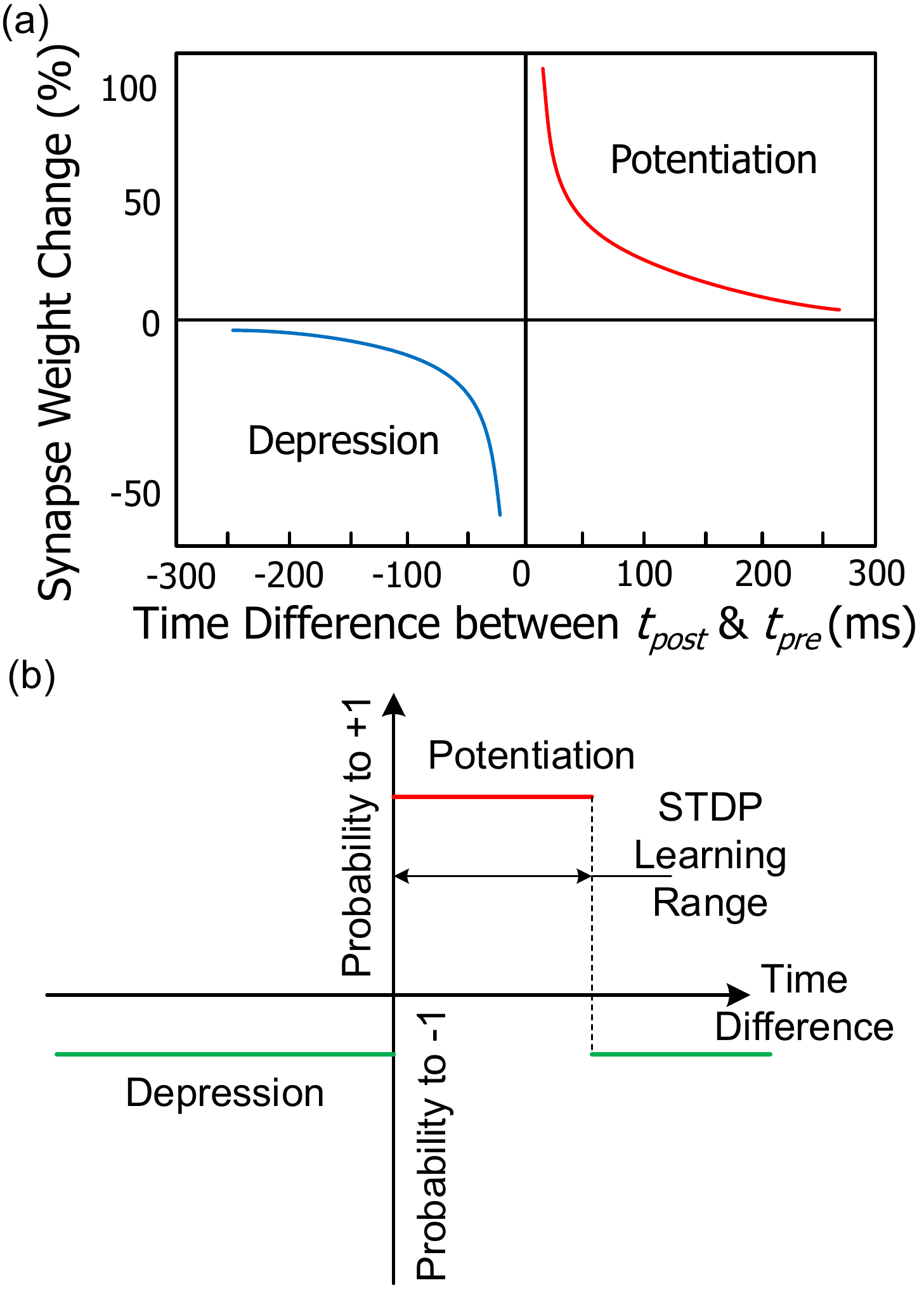}
	\caption{(a) STDP learning curve. (b) Modified STDP learning curve.}
	\label{fig:fig4}
\end{figure}

\subsection{Backpropagation, Learning, and STDP}

\begin{figure*}[t!]
	\centering
	\includegraphics[width=0.85\textwidth]{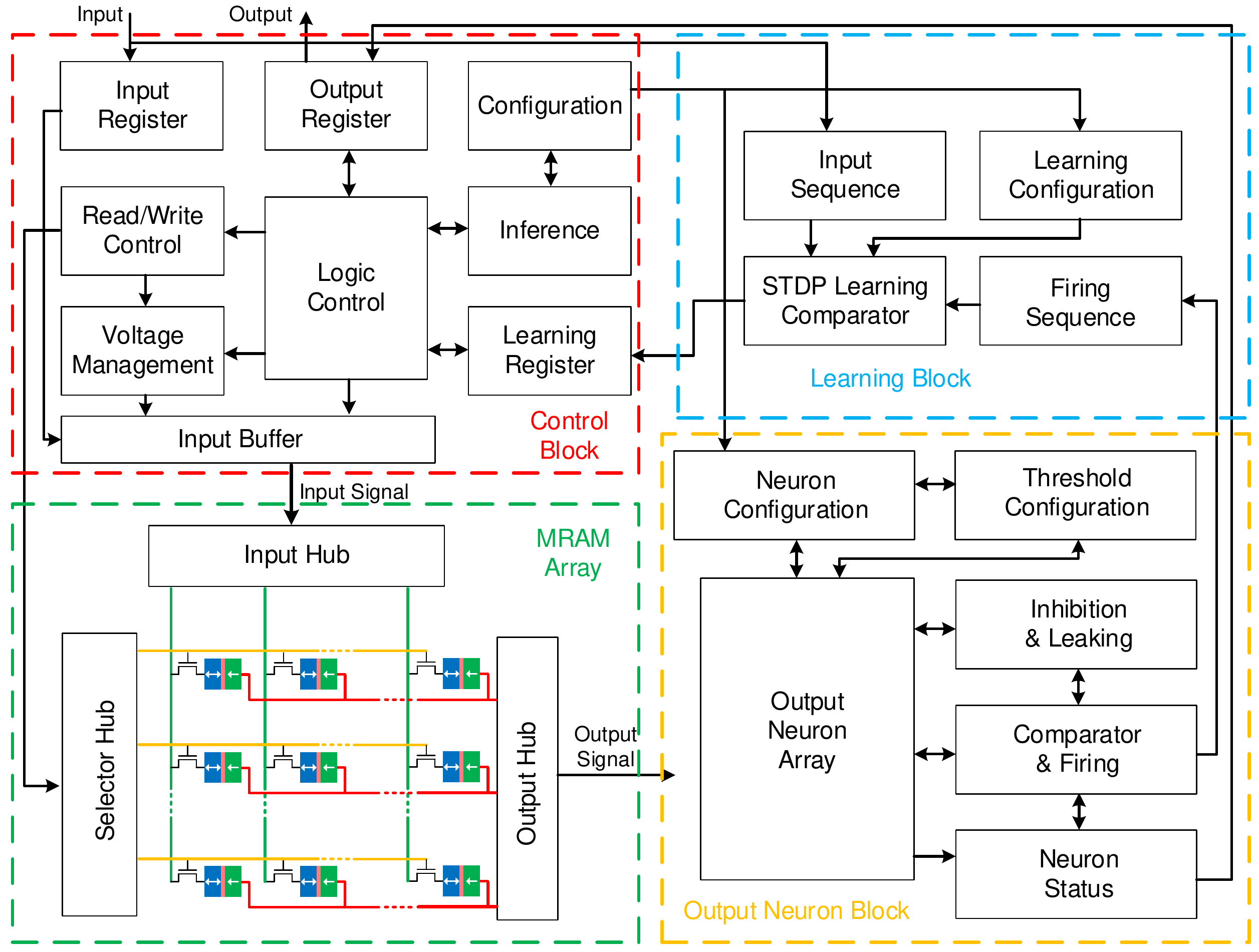}
	\caption{Block diagram of STT-MRAM on-chip online learning and recognition system..}
	\label{fig:fig5}
\end{figure*}

Backpropagation is a widely used supervised learning algorithm for many types of neural networks. Based on the difference between the correct output and the output of the feed-forward recognition, the backpropagation algorithm can update the weights of the synapses. This supervised learning algorithm requires two-way signal propagation and complex mathematical operations that are difficult to efficiently implement in a circuit.

Online learning involves the continual updating of synapse weights while performing recognition operations. In contrast, offline learning systems compute the synapse weights based on the entire dataset prior to deployment for recognition. Online learning thus provides the exciting capability of continually updating synapse weights after deployment, thereby improving recognition accuracy by adapting to new data and environmental conditions.

STDP is a biological process whereby the synaptic connectivity between neurons is updated over time in response to new input information \cite{caporale2008spike,xiang2020computing,liu2020unsupervised}. As illustrated in Fig. \ref{fig:fig4}(a), the STDP learning rule updates the connectivity/weights of the synapses based on the timing information of pulses between the pre-synaptic and post-synaptic neurons. When the post-synaptic neuron fires after the pre-synaptic neuron spike, the weight of the connecting synapse increases; if the post-synaptic neuron fires before the pre-synaptic neuron spike, the weight of this synapse decreases. Through this unsupervised STDP learning rule, synapse training requires only the local information from the neurons connected to each synapse, enabling learning circuits that are much simpler and more energy-efficient than necessary for supervised learning algorithms such as backpropagation that require global information regarding the entire neural network.

\begin{figure}[h]
	\centering
	\includegraphics[width=1\columnwidth]{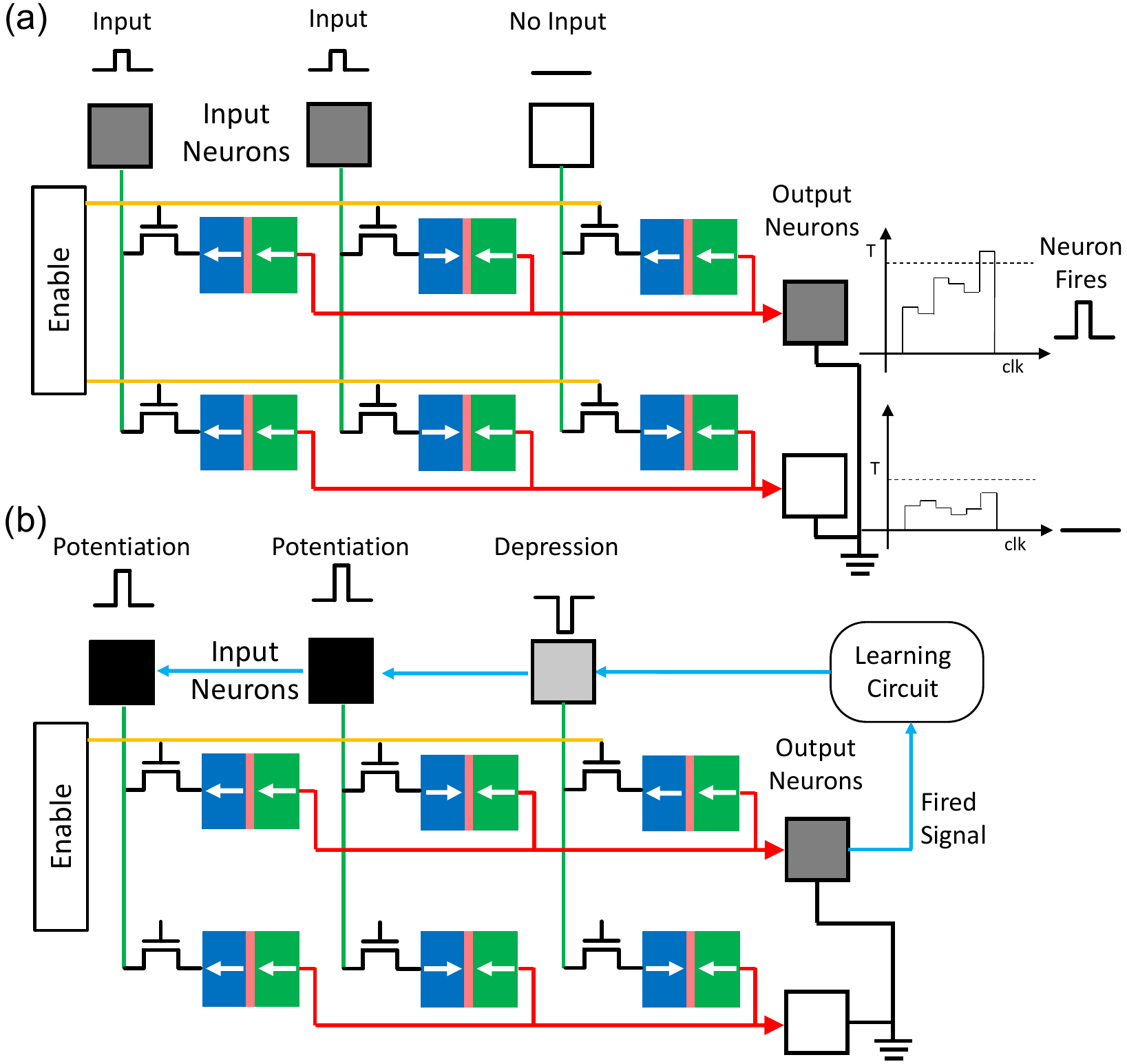}
	\caption{(a) Input neurons provide voltage pulses that flow through MTJ synapses and are integrated by the output neurons. The output neurons fire when their integrated signal is greater than the neuron threshold. (b) During the learning process, a neuron firing is processed by a digital learning circuit to generate potentiation and depression pulses through the synapses connected to the fired neuron. These pulses probabilistically switch the MTJ synapses through stochastic STT-MRAM switching.}
	\label{fig:fig6}
\end{figure}

\subsection{Simplified STDP Learning with Non-Volatile Memory}

To realize STDP learning in practical applications with non-volatile memory devices, Querlioz \textit{et al.} proposed simplified STDP learning rules in which there is an analog change in conductivity \cite{querlioz2012bioinspired} or a probability of binary switching \cite{vincent2015spin}. These changes are discretized as in Fig. \ref{fig:fig4}(b), drastically simplifying the learning circuit implementation. When the output neuron fires within the potentiation time range of the stochastic learning rule, the synapse switches to a low resistance state with a particular potentiation probability. When the input or output neurons fire outside of this potentiation time range, the synapse switches to a high resistance state with a particular depression probability.

\cite{vincent2015spin} \textit{et al.} simulated an MRAM SNN in which this simplified STDP learning was applied to the stochastic switching of STT-MRAM. Their results indicate that even with large synaptic variability, this learning rule can achieve high recognition accuracy on complex recognition task. This simplified STDP learning rule is therefore highly promising for energy-efficient neuromorphic computing with non-volatile memory devices.

The approach of \cite{querlioz2012bioinspired} and \cite{vincent2015spin}, however, cannot be directly applied to the design and fabrication of a learning and recognition circuit. It is not clear how the pre-synaptic and post-synaptic neuron spikes are supposed to generate the learning patterns applied to the MRAM synapses. In particular, the system of \cite{vincent2015spin} appears to be asynchronous, yet provides specific learning patterns dependent on the neuronal spiking activity. In section III, we therefore propose practical learning circuits and a complete synchronous neuromorphic system that can be readily fabricated.

\section{Synchronous STDP Learning Circuit \& System}
The proposed synchronous circuit has four primary blocks, as shown in Fig. \ref{fig:fig5}: control, MRAM array, output neuron, and STDP learning. These four blocks design together comprise a complete neural network that can be directly implemented in a circuit.

\subsection{Control Block}
The control block controls and configures all of the other blocks, and is also the input/output interface to the system. The control block receives a binary input dataset and converts these inputs to pulse signals for the MRAM array, while also providing this input data to the STDP learning block for use in synapse training. Additionally, the system parameters can be configured within the control block to modify the MRAM recognition and learning rules for varying neuromorphic computing tasks. Finally, the control block can also output the SNN status.

\begin{figure*}[!t]
	\centering
	\includegraphics[width=0.85\textwidth]{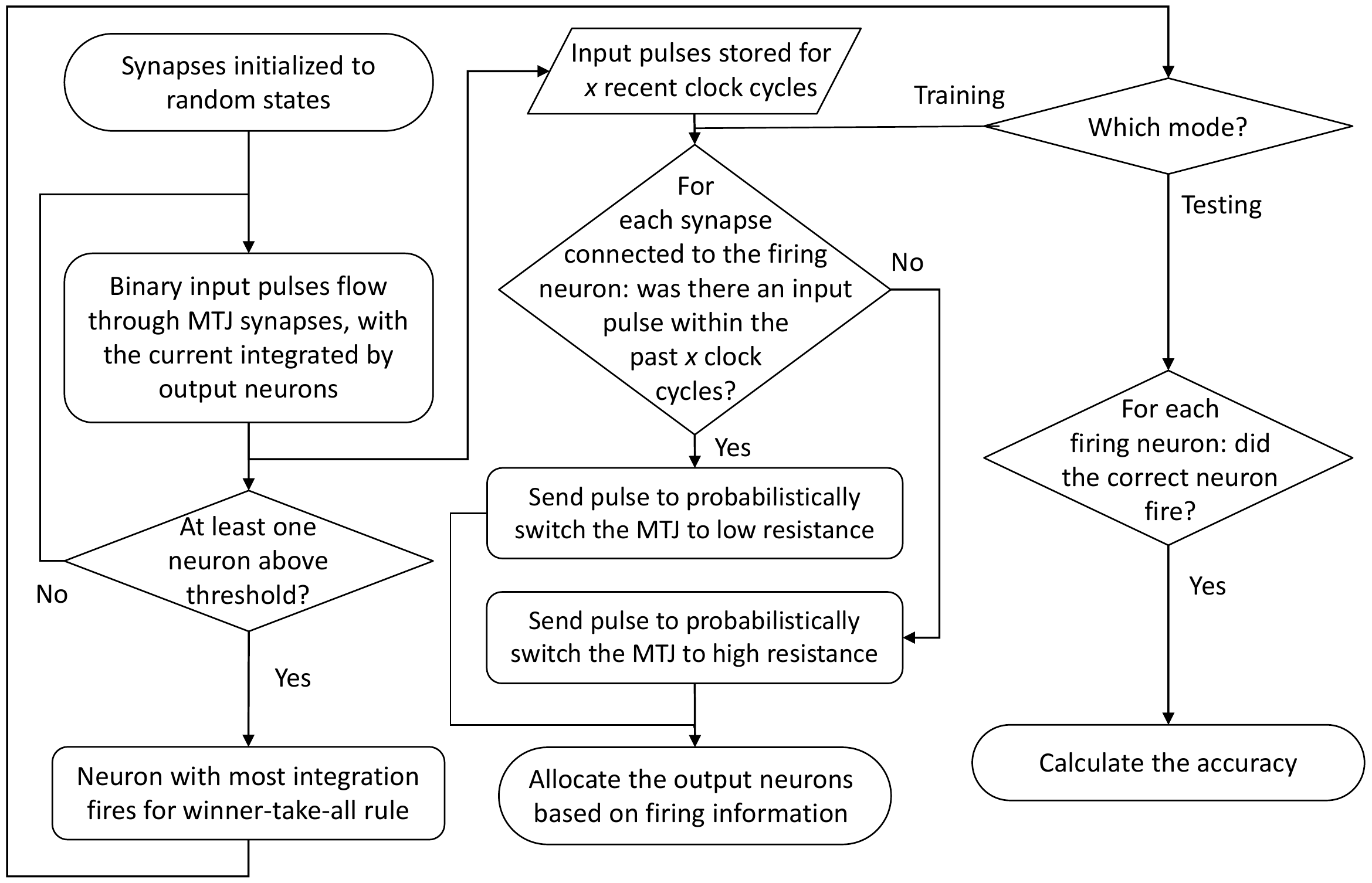}
	\caption{Flow chart of the STT-MRAM learning and recognition algorithms.}
	\label{fig:fig7}
\end{figure*}

\subsection{MRAM Array}
The MRAM array uses a 1T1R configuration for precise control and sneak path reduction. The MTJ cells are arranged in a crossbar to enable the direct computation of vector-matrix multiplication. The input pulses arrive column-wise from the control block, and cause output currents to be produced for each row as a function of the input signals and the MTJ resistances.

\subsection{Output Neuron Block}

The output neuron block contains a column of output neurons, where each neuron is connected to one row of MRAM synapses. Each neuron integrates the current output by a row in the MRAM array, which is compared to a threshold defined separately for each output neuron. The threshold of each neuron is determined by its recent firing activity as in Algorithm \ref{alg0}, with recently active neurons having a high threshold while recently inactive neurons have a low threshold. When an integrated neuron signal is greater than its threshold, the output neuron generates a firing signal. If multiple neuron signals are greater than their thresholds, the winner-take-all circuit will cause a firing signal to be generated by only the neuron with highest ratio of integrated signal to threshold according to Algorithm \ref{alg1}. Whenever an output neuron fires, all output neurons are reset to a state with zero integrated signal, while the firing signal propagates to the STDP learning block. Leaking is continually performed, with each integrated signal reduced toward zero by a particular leaking value during every clock cycle.

\subsection{STDP Learning Block}

The memory in the STDP learning block saves the input dataset from recent clock cycles. Whenever an output neuron fires, the STDP learning block generates a learning signal that is applied to each synapse connected to the neuron that fired (\textit{i.e.}, in the same row). This learning signal is created based on the input sequence stored in the memory: if an input pulse had been received in a particular column since the previous neuron firing, a potentiation signal will be generated; otherwise, a depression signal will be generated. These learning signals are sent to the control block to generate the learning pulses that switch the MTJs between the parallel and anti-parallel states with particular potentiation and depression probabilities. The control block then causes the learning pulses to flow to the synapses connected to the fired output neuron.

\begin{algorithm}[t]
	\caption{Output Neuron Threshold Adjustment} 
    \label{alg0}
	\begin{algorithmic}[1]
	    \FOR {Each clock cycle}
            \FOR{Each output neuron $\mathbf{O}_{j}$}
                \IF{Fired within ${N}_{active}$ clock cycles}
                    \STATE Adjust $\mathbf{O}_{j}$ to high threshold ${T}_{high}$
                \ENDIF  
                \IF{Not fired within ${N}_{inactive}$ clock cycles}
                    \STATE Adjust $\mathbf{O}_{j}$ to low threshold ${T}_{low}$
                \ENDIF
            \ENDFOR
    	\ENDFOR
	\end{algorithmic} 
\end{algorithm}

\begin{algorithm}[t]
	\caption{Output Neuron Firing} 
    \label{alg1}
	\begin{algorithmic}[1]
	    \FOR {Each training dataset image $\mathbf{D}_{train}$ }
            \STATE Continually input this image until at least one output neuron has integration $\geq$ threshold
            \FOR {Each output neuron}
    	        \IF {Integration $\geq$ threshold of this output neuron}
        	        \STATE Add this neuron to the waiting list
    	        \ENDIF
    	    \ENDFOR
            \STATE Calculate the ratio between integration and\\ the threshold of all neurons in the waiting list
            \RETURN The neuron with the maximum ratio
    	\ENDFOR
	\end{algorithmic} 
\end{algorithm}

\subsection{Synchronous Design}

The proposed on-chip unsupervised online learning system is centered on the STT-MRAM shown in Fig. \ref{fig:fig6}, and relies heavily on the synchronous clock. During each clock cycle, the control block generates the input pulses to the MRAM array, and the output neuron block integrates the current from the MRAM array. Later in the same clock cycle, the output neurons are compared to their thresholds to determine whether or not to fire. If a neuron fires, a learning pulse is generated during the next clock cycle and the neuron integrated signals are reset to zero; if no neuron fires, then leaking is performed during the next clock cycle. This synchronous circuit design has been demonstrated via behavioral simulation and can be readily translated to a digital circuit design that can be fabricated and experimental demonstrated.

\section{Online Learning \& Recognition Results}
Behavioral simulations were performed using this STT-MRAM system on the MNIST handwritten digit dataset\cite{lecun1998gradient}. The single layer SNN is shown to achieve a 90\% inference accuracy with the MNIST dataset, proving the functionality of the proposed stochastic binary STT-MRAM SNN system.

\subsection{System Configuration}

\begin{algorithm}[t]
	\caption{Output Neuron Allocation} 
    \label{alg2}
	\begin{algorithmic}[1]
	    \STATE After training is complete
	    \FOR {Each training dataset image $\mathbf{D}_{train}$ }
	        \STATE Continually input this image until an output neuron fires
	        \STATE Output map $\mathbf{M}_{i,j}$ += 1, where $i$ represents the label of the dataset and $j$ represents the label of the fired output neuron
	    \ENDFOR
	    \STATE Sort all $\mathbf{M}_{i,j}$ from high to low
	    \WHILE{Size of output map $\mathbf{M}$ $\neq$ 0}
	        \STATE Select the largest $\mathbf{M}_{i,j}$ in the output map $\mathbf{M}$
	        \STATE Add output neuron $\mathbf{O}_{j}$ to the allocation list $\mathbf{L}_{i}$
	        \STATE Erase all elements of $\mathbf{M}$ with output neuron label $j$
	    \ENDWHILE
	    \RETURN The allocation list $\mathbf{L}$
	\end{algorithmic} 
\end{algorithm}

The system-level simulations were performed in C++ on a binarized MNIST handwritten digit dataset in which the gray-scale pixel data is compared to a threshold. The MTJ tunnel magnetoresistance value is chosen as 300\%\cite{zhang2018addressing,ikeda2008tunnel,song2018demonstration}. The binarized 28x28 input dataset is provided to the 784 input neurons, and varying quantities of output neurons were used (1,000, 2,000, 4,000, 6,000, 8,000, and 10,000). In some network configurations, several synapses in the same row can connect to the same input neurons to compensate for the MRAM synapse stochasticity; this quantity is referred to as \textit{r} (values of 1, 2, 4, and 8 were considered), and these synapses all receive the same STDP learning pulses.

\begin{figure}[!t]
	\centering
	\includegraphics[width=0.95\columnwidth]{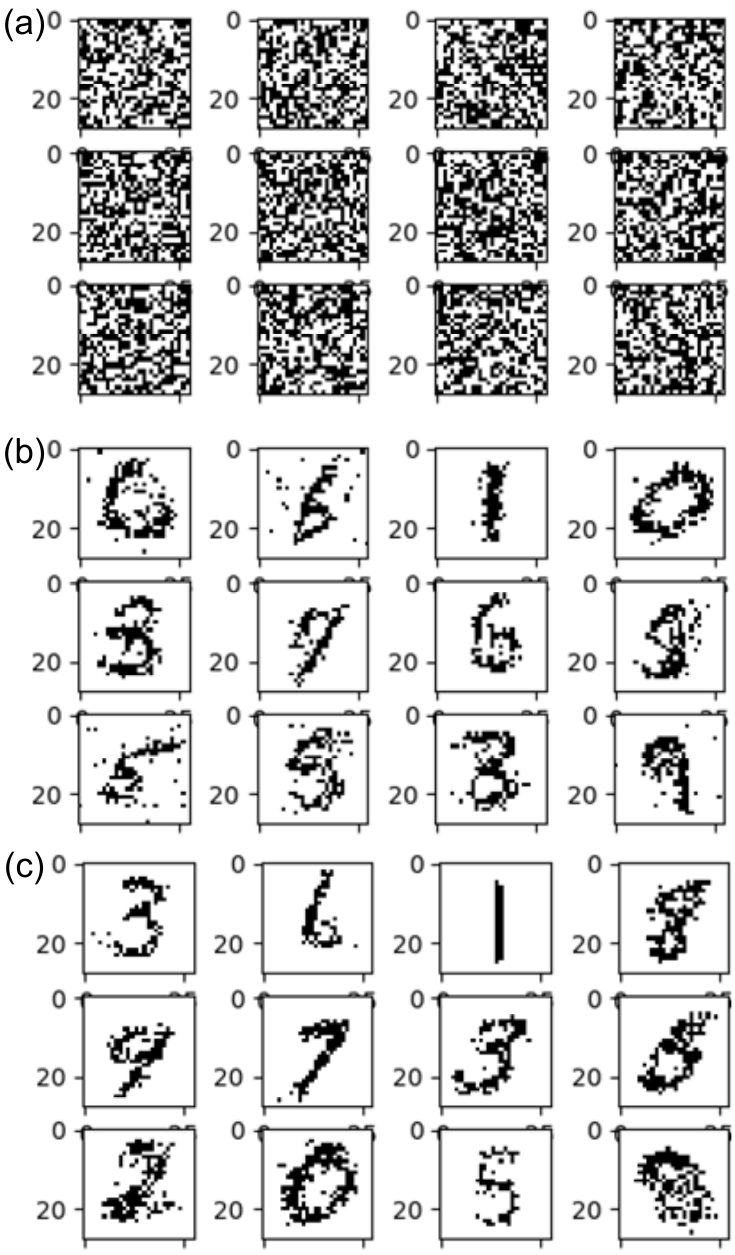}
	\caption{(a) All the STT-MRAM synapses are initialized to random states before learning. (b) The synapse states of selected neurons after training a 500-neuron network with 10,000 MNIST handwritten digits. (c) The states of the same neurons after training the network with 60,000 MNIST handwritten digits. Note that several neurons have specialized in different digits than in (b).}
	\label{fig:fig8}
\end{figure}

The simulation methodology is illustrated in Fig. \ref{fig:fig7}. First, all synapses are initialized to random states. The input then begins to continually generate binary voltages representing the binarized MNIST data, and the resulting currents are integrated by the output neurons. Each input is presented until an output neuron integrates current above its threshold, causing the neuron with the greatest integration ratio to fire. While the 60,000 training images are being provided, the binary synapses are probabilistically switched according to the learning rule whenever an output neuron fires; after the training images have been processed, the output neurons are allocated to particular digits based on their firing activity as shown in Algorithm \ref{alg2}. The recognition accuracy is then calculated based on the percentage of the 10,000 testing images that are correctly labeled by the allocated neurons.

\begin{figure}[!t]
	\centering
	\includegraphics[width=0.95\columnwidth]{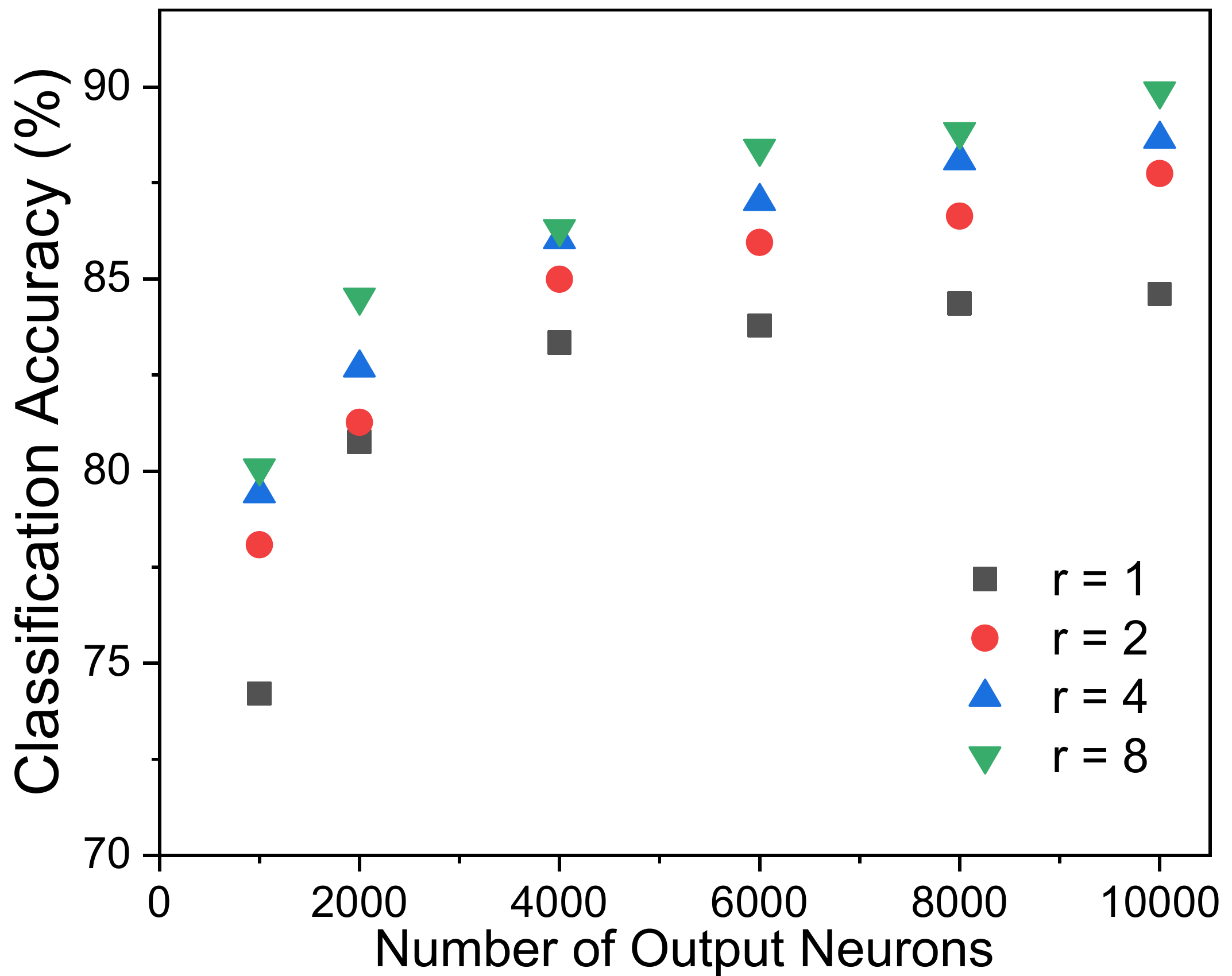}
	\caption{Accuracy as a function of the number of output neurons, where r is the number of synapses representing each pixel.}
	\label{fig:fig9}
\end{figure}

\subsection{MNIST Results and Discussion}

Fig. \ref{fig:fig8} shows the evolution of MRAM synapse weights during the simulation as a result of the STDP learning rule. As can be seen in the figure, some neurons begin to specialize in one digit, but later evolve their specialization to another digit in response to the firing activity of other output neurons.

The inference accuracy results for a single layer MRAM SNN are shown in Fig. \ref{fig:fig9} for varying quantities of output neurons and synapses for each pixel. The inference accuracy clearly increases with both increasing numbers of output neurons and increasing numbers of synapses sharing each pixel. When there are 10,000 output neurons and eight synapses for each pixel, the inference accuracy can reach 90\%. These results are comparable to simulations of unsupervised single layer SNNs based on multilevel memristor evaluated with a similar size and methodology\cite{rathi2018stdp}, demonstrating the feasibility of the proposed binary MRAM approach.

However, multilevel and analog behavioral simulations such as \cite{rathi2018stdp} do not consider the challenges for precisely writing and storing memristor resistance states. Given that the proposed use of MRAM achieves similar inference accuracies in optimistic simulations, it is expected that due to the imprecise writing and drift of analog/multilevel resistance states in memristor and PCM devices, the proposed binary MRAM with stochastic writing will soon be experimentally proven to provide higher accuracies than can be achieved with memristors and PCM.

\section{Conclusion}
This work proposes a neural network with synchronous STDP learning based on stochastic STT-MRAM switching. A single layer SNN is shown to achieve a 90\% inference accuracy with the MNIST dataset, proving the ability of stochastic binary STT-MRAM switching to emulate analog synapse resistances. The synchronous design of the circuit can be easily migrated to digital circuit design written in a hardware language design and synthesized for tape-out. The proposed synchronous learning with binary stochastic MRAM based SNN is therefore a promising and practical solution for energy-efficient neuromorphic computing.


%



\section*{Acknowledgment}
This work is supported by Semiconductor Research Corporation (SRC) Task No. 2810.030 through UT Dallas’ Texas Analog Center of Excellence (TxACE).

\ifCLASSOPTIONcaptionsoff
  \newpage
\fi

\IEEEtriggeratref{10}


\bibliographystyle{IEEEtran}
\bibliography{MRAMbib}

%








\end{document}